\documentclass[11pt]{article}

\usepackage[preprint]{acl}

\usepackage{times}
\usepackage{latexsym}

\usepackage[T1]{fontenc}

\usepackage[utf8]{inputenc}

\usepackage{microtype}

\usepackage{inconsolata}

\usepackage{multicol}
\usepackage{multirow}
\usepackage{graphicx}
\usepackage{xcolor,soul}
\usepackage{CJKutf8}
\usepackage{tabularx}
\usepackage{booktabs}
\usepackage{amsmath}
\usepackage{makecell}
\usepackage{subcaption}
\usepackage[misc]{ifsym}
\usepackage{amssymb}
\usepackage{pifont}
\usepackage[export]{adjustbox}

\usepackage{textcomp}  
\usepackage{scalerel}  

\definecolor{amsblue}{RGB}{0,72,120}

\newcommand{\red}[1]{\textcolor{red}{#1}}
\newcommand{\blue}[1]{\textcolor{blue}{#1}}

\newcommand{\pmt}[1]{$_{\pm\text{#1}}$}

\def\tickemoji {\scalerel*{\includegraphics{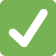}}{\textrm{\textbigcircle}}}
\def\crossemoji {\scalerel*{\includegraphics{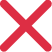}}{\textrm{\textbigcircle}}}

\newcommand{\zhsmall}[1]{\begin{CJK*}{UTF8}{gbsn}\small{#1}\end{CJK*}}

\def\hindiworda {\scalerel*{\includegraphics{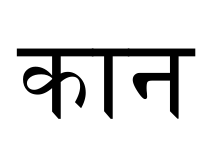}}{\textrm{\textbigcircle}}}
\def\koreanwordb {\scalerel*{\includegraphics{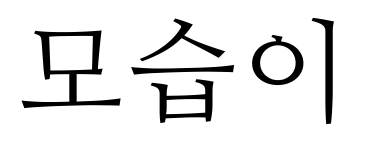}}{\textrm{\textbigcircle}}}

\usepackage{tcolorbox}
\tcbuselibrary{breakable} 
\usepackage{tikz}
\usetikzlibrary{positioning,fit}

\newtcolorbox{promptbox}[2][]{
  breakable,
  colback=gray!3,
  colframe=amsblue,
  boxrule=0.8pt,
  arc=2mm,
  left=6pt,
  right=6pt,
  top=6pt,
  bottom=6pt,
  fonttitle=\bfseries,
  title={#2},
  #1
}

\title{When Rules Learn: A Self-Evolving Agent for Legal Case Retrieval}

\author{
\textbf{Mingxu Tao\textsuperscript{1,2}},
\textbf{Jiawei Hu\textsuperscript{1,2,3}},
\textbf{Xian Zhou\textsuperscript{1,2}},
\textbf{Wenpeng Hu\textsuperscript{1,2}},
\textbf{Jiajun Cheng\textsuperscript{1,2}} \\
\textbf{Yunbo Cao\textsuperscript{1,2\,\Letter}\thanks{Corresponding author.}},
\textbf{Zhunchen Luo\textsuperscript{1,2\,\Letter}\footnotemark[1]},
\textbf{Guotong Geng\textsuperscript{1,2}}\\
\textsuperscript{1}Center of Information Research, AMS\\
\textsuperscript{2}Discipline and Technology Research Center for Large Model Intelligence Applications\\
\textsuperscript{3}Hebei University of Engineering\\
  \texttt{\{thomas.mx.tao,zhunchenluo\}@gmail.com},\ \texttt{caoyunbo@hotmail.com}\\}

\begin{document}
\maketitle
\begin{abstract}

Legal case retrieval remains challenging due to the complexity of legal language and the need for precise lexical alignment between queries and relevant cases. 
Although dense retrieval models have achieved notable progress, empirical studies show that BM25 continues to serve as a strong baseline in this domain. It motivates us to propose a self-evolving framework for rule-driven query rewriting that enhances BM25 without any parameter training. The framework equips an LLM-based agent with an automatic evaluation environment, enabling it to iteratively \textbf{create rewriting rules}, \textbf{plan validation experiments} over rule combinations, and \textbf{eliminate ineffective rules} based on historical feedbacks. 
We evaluate our method on the Chinese legal case retrieval benchmark LeCaRD-v2. 
Experimental results demonstrate that the proposed framework outperforms non-evolutionary baselines, including human-designed rules and greedy rule selection, particularly when powered by a high-capacity core LLM. 
We also conduct detailed analyses to investigate the mechanisms underlying self-evolution. Our findings reveal that LLM's capabilities to leverage previous experimental results and its intrinsic knowledge of rule elimination play critical roles in refining the rule set via self-evolution.

\end{abstract}

\section{Introduction}

Legal case retrieval plays a vital role in supporting judicial decision-making, legal consultation, and a wide range of downstream legal applications. 
Given a natural-language description of disputes, retrieval systems are expected to identify relevant prior cases that share similar facts, legal issues, or applicable statutes.
This task remains challenging due to the complexity of legal language, the length of legal documents, and the requirement for precise matching of legal facts, statutes, and judicial reasoning.

Despite the rapid progress of dense retrieval methods~\cite{hu-etal-2024-ella,su2024caseformerpretraininglegalcase} based on neural embeddings~\cite{chen-etal-2024-m3,sailer-li-et-al-2023,tang2025reakase8blegalcaseretrieval}, their effectiveness in legal case retrieval remains limited. 
Previous works~\cite{rosa2021yesbm25strongbaseline,deng-etal-2024-element} reveal that lexical matching methods continue to serve as strong and competitive baselines in the legal case retrieval task.
Our empirical study on LeCaRD-v2~\cite{levardv2-li-et-al-2024} also shows that BM25~\cite{bm25} consistently outperforms several representative dense retrieval models.
These findings highlight that, rather than replacing BM25 with dense retrievers, a more promising direction is to enhance BM25 by lexically aligning queries with the relevant documents.

Building upon the strengths of BM25, we focus on improving retrieval performance through \textbf{query rewriting}.
Query rewriting aims to bridge the gap between queries and relevant legal cases by enriching queries with legal terminology, synonymous expressions, or alternative formulations.
Recent progress in large language models (LLMs)~\cite{yang2025qwen3technicalreport,openai2025gptoss120bgptoss20bmodel,deepseekai2025deepseekr1incentivizingreasoningcapability,gemmateam2025gemma3technicalreport} make it feasible to perform rule-driven query rewriting, where an LLM rewrites queries by following explicitly rewriting rules.
Such rules offer interpretability and controllability, which are particularly desirable in the legal domain.
However, designing high-quality rewriting rules typically requires substantial domain expertise, while naive rule generation may lead to suboptimal performance.

To address these limitations, we propose a \textbf{self-evolution framework} for the rule-driven query rewriting.
The framework equips an LLM-based agent with an automated validation environment that enables iterative optimization of the rule set without any parameter updates.
Specifically, the agent can autonomously (1) create new query-rewriting rules, (2) plan validation experiments by selecting combinations of rules, and (3) eliminates ineffective rules based on experimental history.
Through repeated interaction with the environment, the agent continuously refines its rule set, improving the effectiveness of retrieval in a training-free manner.

We evaluate the proposed framework on the LeCaRD-v2 benchmark~\cite{levardv2-li-et-al-2024}.
Experimental results show that the self-evolution framework can outperform the non-evolutionary baselines, including human-designed rules and the greedy strategy for rule selection.
We also conduct further analyses to investigate the mechanisms underlying self-evolution, revealing that the effectiveness of self-evolution depends on the LLM’s ability to plan new experiments by leveraging historical results, as well as its capability to determine appropriate timing for rule elimination.

In summary, our contributions are as follows:
\begin{itemize}
    \item We revisit legal case retrieval, demonstrating that BM25 still remains a strong baseline, and enhance it with rule-driven query rewriting strategies.
    \item We propose a novel self-evolution framework, which enables the agent to iteratively optimize rewriting rules through rule generation, experiment planning, and rule elimination, without any gradient-based training.
    \item We conduct extensive investigations to characterize the behavioral dynamics of self-evolving LLM agents, shedding light on the intrinsic factors affecting agent’s capacity for self-evolution.
\end{itemize}

\section{Related Works}

Prior work on autonomous or self-evolving agent systems has recently garnered increasing attention, particularly in contexts requiring adaptive behavior and iterative improvement. 

Several studies have explored mechanisms by which agents can refine their internal strategies without heavy human supervision. For instance, AgentEvolver~\cite{zhai2025agentevolverefficientselfevolvingagent} introduces self-questioning, experience-guided exploration, and fine-grained credit attribution to enable agents to generate their own tasks, reuse experiences, and improve exploration efficiency in complex environments. The process can reduce reliance on handcrafted datasets and fixed reinforcement learning pipelines, extending broader paradigms of self-evolving agent architectures. 
EvolveSearch~\cite{zhang-etal-2025-evolvesearch} proposes iterative self-evolving search agents. The system refines its retrieval behavior via ongoing self-improvement loops, indicating that iterative evolution paradigms can yield measurable gains for information search tasks. 
However, these methods depend on reinforcement learning to optimize the model’s action selection strategy. They face challenges when applied to the legal case retrieval task due to limited training data.

Recent work~\cite{suzgun2025dynamiccheatsheettesttimelearning,zhang2025agenticcontextengineeringevolving} also focuses on evolving the contextual information inputted into LLMs. They treat the contexts in prompts as dynamic entities that can be incrementally refined.
The previous method, Agentic Context Engineering~\cite{zhang2025agenticcontextengineeringevolving}, proposes to generate, reflect on, and select contextual elements in a \textit{playbook} memory.
These methods emphasize evolving inputs and contextual representations to improve downstream performance, which resonates with our framework of evolving rule sets for query rewriting rather than training model parameters. 
However, our work explicitly evolves a set of structural rules, which serve as the instructions for downstream query rewriting, rather than the agent's intrinsic prompts.

\section{Preliminaries}

Legal case retrieval poses unique challenges due to the complexity of legal wordings, long document lengths, and the need for precise lexical matching of legal facts, statutes, and judicial reasoning.
Despite the rapid advancement of dense retrieval methods based on neural embeddings, their effectiveness in legal case retrieval remains limited.
Previous work~\cite{rosa2021yesbm25strongbaseline} argues that the BM25 function~\cite{bm25} continues to serve as a strong baseline. 

We compare BM25 with several representative dense retrieval models on the held-out test set of LeCaRD-v2~\cite{levardv2-li-et-al-2024}, a Chinese legal case retrieval benchmark.
We consider three models: \texttt{bge-m3}~\cite{chen-etal-2024-m3}, \texttt{SAILER\_zh}~\cite{sailer-li-et-al-2023}, and \texttt{ReaKase-8B}~\cite{tang2025reakase8blegalcaseretrieval}. 
Among these models, \texttt{bge-m3} is a general embedding model for dense retrieval, while the other two are embedding models that have been continually trained on legal-domain data.


Table~\ref{tab:preliminary_scores} presents the results under different cutoffs.
BM25 achieves the highest average recall across different scales of retrieved candidates, outperforming both general and legal-domain embedding models.
Although \texttt{ReaKase-8B} and \texttt{SAILER\_zh} demonstrate competitive performance in Recall@1000, their effectiveness remains significantly inferior to BM25 when limiting the amount of retrieved candidates. 
Compared with BM25, the two models exhibit performance drops of 7.60\%\textasciitilde 19.07\% on Recall@50 and 6.74\%\textasciitilde 20.51\% on Recall@100.

\begin{table}[t]
\small
\tabcolsep=0.40em
\centering
\begin{tabular}{lccccccc}
\toprule
\multirow{2}{*}{\textbf{Method}} & \multirow{2}{*}{$\mathcal{L}$} & \multicolumn{5}{c}{\textbf{Recall@\texttt{k}}} & \multirow{2}{*}{\textbf{Avg.}}\\
\cmidrule{3-7}
 & & \texttt{k=50} & \texttt{100} & \texttt{200} & \texttt{500} & \texttt{1000} & \\
\midrule

BM25 & \crossemoji & 
    \textbf{38.29} & \textbf{48.79} & \textbf{58.89} & 70.14 & 77.32 & \textbf{58.69}\\
\texttt{bge-m3} & \crossemoji &
    27.88 & 37.05 & 47.51 & 63.33 & 74.45 & 50.04 \\
\texttt{SAILER} & \tickemoji &
    19.22 & 28.28 & 41.18 & 63.45 & 78.93 & 46.21 \\
\texttt{ReaKase} & \tickemoji &
    30.69 & 42.05 & 54.82 & \textbf{73.32} & \textbf{84.47} & 57.07 \\
\bottomrule
\end{tabular}
\caption{Recall scores for BM25 and dense retrieval based on distinct embedding models on LeCaRD-v2. $\mathcal{L}$ denotes whether the model is trained on legal data.}
\label{tab:preliminary_scores}
\end{table}

The strong performance of BM25 motivates us to enhance it through query rewriting rather than replacing it with dense retrieval models.
Query rewriting aims to bridge the gap between user's queries and relevant legal cases by enriching or reformulating queries with legal terms.
We leverage a reasoning LLM to follow given rules and rewrite queries. Figure~\ref{fig:example_of_rule} shows an example of the rules. See the prompt of query rewriting in Appendix~\ref{app:prompt_rewriting}.

\begin{figure}[h]
    \centering
    \begin{tcolorbox}[arc=1mm,boxrule=1pt,left=4pt,right=4pt,top=4pt,bottom=4pt]
    \small
    \textit{The original rule in Chinese:}
    
    \zhsmall{\textbf{同义词和术语变体扩展：}对于每个关键法律概念，添加其同义词、近义词或常见变体。}

    \zhsmall{\textbf{格式：}[术语1]、[术语2]、[术语3]、……}
    \end{tcolorbox}
    
    \begin{tcolorbox}[arc=1mm,boxrule=1pt,left=4pt,right=4pt,top=4pt,bottom=4pt]
    \small
    \textit{The translated rule in English:}
    
    \textbf{Synonym and Terminology Variant Expansion:} For each key legal concept, augment the query by incorporating its synonyms, semantically related terms, or commonly used variants.
    
    \textbf{Format:} [Term A], [Term B], [Term C], ...
    \end{tcolorbox}
    \caption{An example of query-rewriting rules.}
    \label{fig:example_of_rule}
\end{figure}

\begin{figure*}[t]
\centering
\includegraphics[width=1\textwidth]{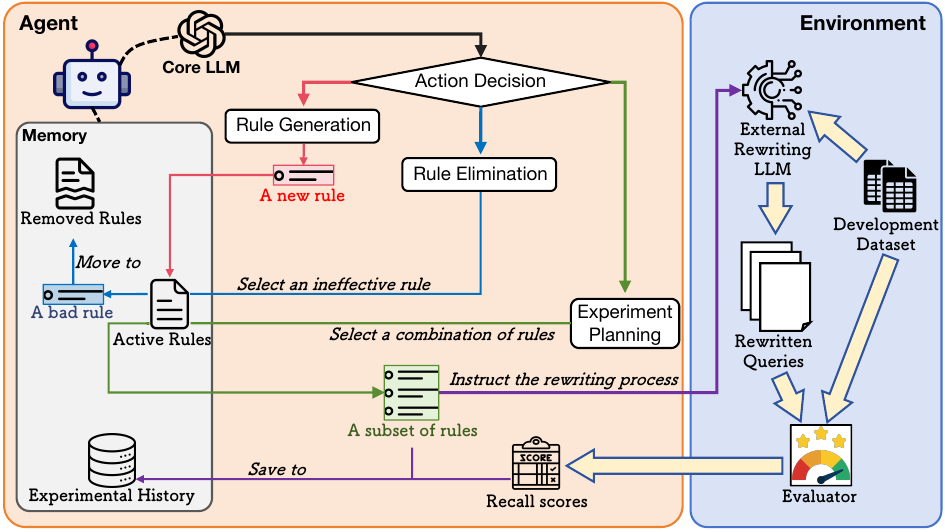}
\caption{Self-evolution based on rule generation, experiment planning, and rule elimination.}
\label{fig:method}
\end{figure*}

\section{Self-evolution Framework}

We propose a self-evolution framework that enables an LLM-based agent to autonomously discover, examine, and refine query-rewriting rules for legal case retrieval. The framework is a closed-loop agent-environment system, where adaptation emerges from iterative interaction, rather than gradient-based optimization.
Figure~\ref{fig:method} illustrates the self-evolution framework.

\subsection{Action Decision}

The agent selects actions to drive the self-evolution process.
The action space consists of three categories: (1) \textbf{creating a new rule}, (2) \textbf{planning an experiment}, and (3) \textbf{eliminating an ineffective rule}. 

At the interaction step $t$, the agent selects an action based on its internal memories, which comprises a sequence of recent actions $A^{(t)}=\left<a_{t-k},\,\cdots,\,a_{t-1}\right>$ truncated by a fixed length $k$, the current set of rules $R^{(t)}$, and the accumulated experimental results $S^{(t)}$. We denote the action selected at time $t$ as:
$$a_t = \pi(A^{(t)}, R^{(t)}, S^{(t)}),$$
where $\pi$ is the decision policy that is based solely on the prompt in Appendix~\ref{app:prompt_action_decision}.
The policy operates without any parameter updates during the evolution process.

\subsection{Rule Generation}

Creating a new rule enables the agent to expand the search space of query-rewriting strategies. 
At step $t$, the agent generates a candidate rule by jointly analysing the current active rule set $R^{(t)}$, the set of eliminated rules $\bar{R}^{(t)}$, and the accumulated experimental scores $S^{(t)}$. 
This analysis focuses on identifying which rewriting operations contribute positively to recall improvements and diagnosing failure patterns exhibited by removed rules. The agent then generate single new rule:
$$r_{n}=f_{\text{create\_rule}}(R^{(t)}, \bar{R}^{(t)}, S^{(t)}),$$
where $n$ equals to $|R^{(t)}|+|\bar{R}^{(t)}|+1$.

The agent may generate a new rule through two mechanisms. First, it can edit an existing rule by modifying its description of rewriting operation, with the goal of amplifying observed effective behaviors. Second, the agent can propose a novel rule that introduces a completely different rewriting strategy not present in $R^{(t)}$. To prevent degeneration and repeated exploration of unproductive patterns, we require that the generated rule should not be similar to the eliminated ones in $\bar{R}^{(t)}$. 

\subsection{Experiment Planning}
Planning an experiment allows the agent to assess the effectiveness of different combinations of  rules for query rewriting.
At step $t$, the agent selects a subset of active rules from the current rule set $R^{(t)}$, denoted as $C^{(t)} \subseteq R^{(t)}$. 
The environment receives these rules and rewrites the query by invoking an external model.
It then feeds back the test results of rewritten queries $s_{C^{(t)}}$ to agent.

We employ the experiment history $S^{(t)}$ to guide the selection of rule combinations. 
Based on this history, the agent estimates which combinations are likely to yield further recall improvements, such as complementary effects.
To ensure efficient exploration, the agent is restricted from selecting any combination that has already been evaluated in prior experiments.

We denote the experiment planning as:
$$C^{(t)}=f_{\text{plan\_exp}}(R^{(t)}, S^{(t)}).$$ 
The agent then appends the experimental result $\left<C^{(t)}, s_{C^{(t)}}\right>$ to $S^{(t)}$.

\subsection{Rule Elimination}

Eliminating ineffective rules plays a vital role to control the growth of the strategy space, preventing the agent from exploring unproductive combinations of rules. 
However, a rule may bring more improvements in part of strategies but not in others. 
The premature  removal of potentially effective rules can hinder long-term performance.
To mitigate this risk, we adopt a two-stage mechanism with self-consistency~\cite{wang2023selfconsistency}.

At step $t$, the agent first determines whether the rule set $R^{(t)}$ contains any rule that should be eliminated. This decision is repeated independently for $n_1$ trials based on the same memories $\left<R^{(t)}, S^{(t)}\right>$. 
If the agent concludes that at least one rule should be removed in more than $\delta_1 \times n_1$ times, the elimination process proceeds to the second stage; otherwise, no rule is removed at this step.

In the second stage, the agent identifies a specific rule for removal. It selects a candidate rule from $R^{(t)}$ over $n_2$ independent trials, again conditioned on $\left<R^{(t)}, S^{(t)}\right>$. A rule $r \in R^{(t)}$ can be eliminated if the agent selects it in more than $\delta_2 \times n_2$ times. The removed rule is transferred from the active rule set $R^{(t)}$ to the eliminated rule set $\bar{R}^{(t)}$, and is excluded from subsequent experiment planning.

We formulate the elimination process as:
\begin{equation*}
    r_{\text{elim}} =
\begin{cases}
\begin{split}
\arg\max_{r \in R^{(t)}} \text{Count}(r), 
\end{split} & \text{if } m > \delta_2 \times n_2, \\
\varnothing, & \text{otherwise},
\end{cases}
\end{equation*}
where $m$ equals to $\mathbf{\max}_{r \in R^{(t)}} \text{Count}(r)$. We use $\text{Count}(r)$ to denote the number of times in which the rule $r$ is selected for removal across $n_2$ trials.
We believe this two-stage consensus mechanism can mitigate the risk of discarding effective rules due to stochastic.

\subsection{Interactive Environment}

The environment serves as an automatic test bed that executes and evaluates the agent’s decisions. 
The environment can invoke a specific LLM for query rewriting, whose parameters are \textbf{frozen} and does not perform any optimization.

If the agent plans to experiment at step $t$, it yields a subset of rules $C^{(t)} \subseteq R^{(t)}$ to the environment. The environment employs the external LLM to rewrite queries according to the selected rules.
It subsequently evaluates the effectiveness of these refined queries for legal case retrieval, providing the recall scores $s_{C^{(t)}}$ back to the agent.

\section{Experiments}

\subsection{Experimental Setup}

\paragraph{Benchmark} We evaluate the self-evolution framework on the Chinese legal case retrieval benchmark LeCaRD-v2~\cite{levardv2-li-et-al-2024}, which contains 800 queries and a retrieval pool with 55,192 documents. 
We partition the dataset into two sets: 
a {development set} of 100 instances and a {test set} of 700 instances.

\paragraph{Baselines} We compare our framework against three baseline methods to assess the effectiveness of self-evolution.
\label{sec:setup}

(1) \textbf{Human-written rules.} We manually design three rules based on expert knowledge in legal case retrieval, and apply them jointly to query rewriting.

(2) \textbf{LLM-generated rules without evolution~(\textsc{LlmGen}).} We directly employ the core LLM of the agent to generate multiple rules in a single pass, with a human-written rule as examplar. We adopt the combination of generated rules and the human-written rule for rewriting.

(3) \textbf{Greedy Strategy~(\textsc{Greedy}).} Based on the rules generated in Method~(2) across multiple runs, we examine each individual rule on the development set. We select three top-performing rules as a group for rewriting.

\paragraph{Implementation Details} In the environment, we employ \texttt{Qwen3-4B-Thinking}~\cite{yang2025qwen3technicalreport} to rewrite queries in all experiments. We concatenate the rewritten query with the original query, and then use the BM25 algorithm for retrieval. We set $k_1$ to $1.2$ and $b$ to $0.75$. 

In the self-evolution process, we utilise four LLMs as the agent's core model, respectively, including \texttt{Qwen3-4B-Thinking}, \texttt{Qwen3-30B-A3B-Thinking}~\cite{yang2025qwen3technicalreport}, \texttt{gpt-oss-20b}, and \texttt{gpt-oss-120b}~\cite{openai2025gptoss120bgptoss20bmodel}. 
Following the baseline methods, we initially introduce one human-written rule as a seed to the agent’s active rule set $R$.
To ensure self-consistency in rule elimination, we set $n_1$ and $n_2$ to $7$, $\delta_1$ and $\delta_2$ to $0.5$.
We conduct the self-evolution process entirely on the development set. 
The agent explores for up to 500 steps. 
We then select the subset of rules with the best performance from experiment history $S$ and evaluate its effectiveness on the held-out test set.

For each setting, we conduct five independent runs. Given the inherent stochasticity of LLM-based query rewriting, we perform the rewriting process five times and report the average performance.
See more details in Appendix~\ref{app:imple_details}.

\subsection{Main Results}

\begin{table*}[t]
\centering
\begin{tabular}{lcccccc}
\toprule
\multirow{2}{*}{\textbf{Method}} & \multicolumn{5}{c}{\textbf{Recall@k}} & \multirow{2}{*}{\textbf{Avg.}}\\
\cmidrule{2-6}
 & k=50& k=100 & k=200 & k=500 & k=1000 & \\
\midrule
\multicolumn{6}{l}{\textit{Heuristic Baselines}} \\
\textsc{BM25} & 
    38.29 & 48.79 & 58.89 & 70.14 & 77.32 & 58.69 \\
\textsc{Human-1} &
    38.64\pmt{0.17} & 49.37\pmt{0.18} & 59.77\pmt{0.16} & 71.74\pmt{0.13} & 79.07\pmt{0.19} & 59.72 \\
\textsc{Human-3} &
    38.76\pmt{0.03} & 49.91\pmt{0.10} & 60.40\pmt{0.15} & 72.91\pmt{0.16} & 80.63\pmt{0.18} & 60.52 \\
\midrule
\multicolumn{6}{l}{\textit{Core Model: Qwen3-4B-Thinking}} \\
\textsc{LlmGen} & 
   38.41\pmt{0.11} & 49.41\pmt{0.23} & 59.83\pmt{0.35} & 72.05\pmt{0.56} & 79.68\pmt{0.76} & 59.88 \\
\textsc{Greedy} &
   \textbf{38.64}\pmt{0.01} & \textbf{49.61}\pmt{0.07} & 60.05\pmt{0.14} & 71.97\pmt{0.12} & 79.46\pmt{0.13} & 59.95 \\
\textsc{Evolve}~(Ours) &
   38.59\pmt{0.26} & 49.53\pmt{0.35} & \textbf{60.07}\pmt{0.41} & \textbf{72.62}\pmt{0.49} & \textbf{80.66}\pmt{0.74} & \textbf{60.29} \\
\midrule
\multicolumn{6}{l}{\textit{Core Model: Qwen3-30B-A3B-Thinking}} \\
\textsc{LlmGen} & 
   38.40\pmt{0.30} & 49.49\pmt{0.32} & 59.94\pmt{0.31} & 72.17\pmt{0.61} & 79.72\pmt{0.73} & 59.95 \\
\textsc{Greedy} &
   \textbf{38.53}\pmt{0.19} & \textbf{49.87}\pmt{0.22} & \textbf{60.52}\pmt{0.01} & 73.05\pmt{0.13} & 80.88\pmt{0.11} & 60.57 \\
\textsc{Evolve}~(Ours) &
   38.38\pmt{0.19} & 49.61\pmt{0.20} & 60.38\pmt{0.42} & \textbf{73.39}\pmt{0.31} & \textbf{81.75}\pmt{0.41} & \textbf{60.70} \\

\midrule
\multicolumn{6}{l}{\textit{Core Model: gpt-oss-20b}} \\
\textsc{LlmGen} & 
   38.04\pmt{0.46} & 49.17\pmt{0.45} & 59.93\pmt{0.41} & 72.65\pmt{0.22} & 80.70\pmt{0.41} & 60.10 \\
\textsc{Greedy} &
   \textbf{39.42}\pmt{0.30} & \textbf{50.40}\pmt{0.28} & 61.16\pmt{0.14} & 73.55\pmt{0.26} & 81.47\pmt{0.20} & 61.20 \\
\textsc{Evolve}~(Ours) &
   38.99\pmt{0.17} & 50.25\pmt{0.20} & \textbf{61.47}\pmt{0.16} & \textbf{74.04}\pmt{0.33} & \textbf{81.96}\pmt{0.42} & \textbf{61.34} \\

\midrule
\multicolumn{6}{l}{\textit{Core Model: gpt-oss-120b}} \\
\textsc{LlmGen} & 
   38.65\pmt{0.53} & 49.68\pmt{0.54} & 60.24\pmt{0.51} & 72.58\pmt{0.58} & 80.45\pmt{0.74} & 60.32 \\
\textsc{Greedy} &
   39.33\pmt{0.28} & 50.13\pmt{0.23} & 60.81\pmt{0.29} & 73.19\pmt{0.16} & 81.25\pmt{0.15} & 60.94 \\
\textsc{Evolve}~(Ours) &
   \textbf{39.69}\pmt{0.34} & \textbf{50.64}\pmt{0.33} & \textbf{61.22}\pmt{0.16} & \textbf{73.90}\pmt{0.06} & \textbf{82.29}\pmt{0.17} & \textbf{61.55} \\

\bottomrule
\end{tabular}
\caption{Recall scores on the held-out test set of LeCaRD-v2. We report the standard errors across five runs and five times of repeatedly query rewriting.}
\label{tab:main_results}
\end{table*}

Table~\ref{tab:main_results} reports the retrieval recalls on the LeCaRD-v2 held-out test set. 
We include BM25 as a vanilla retrieval baseline, along with two heuristic baselines: \textsc{Human-1}, which applies single human-provided rule for query rewriting, and \textsc{Human-3}, which combines three human-written rules.
Across all cutoff \texttt{k}, \textsc{Human-3} consistently outperforms \textsc{BM25} and \textsc{Human-1}, indicating that a combination of high-quality rules with expert knowledge can effectively improve legal case retrieval.

Beyond heuristic baselines, we evaluate the proposed self-evolution framework using different core LLMs for the agent and compare it against two non-evolutionary baselines introduced in Section~\ref{sec:setup}.
These baselines represent strong alternatives that exploit LLM generation or empirical rule selection without iterative evolution.

When using smaller models with less than 20B parameters as the agent’s core LLM, the self-evolution framework does not yield consistent improvements, particularly at lower cutoffs.
Self-evolution performs worse than \textsc{Greedy} by Recall@50 of 0.05\%\textasciitilde 0.43\% and Recall@100 of 0.08\%\textasciitilde 0.26\%.
The results suggest that smaller models may lack sufficient reasoning capacity to balance performance across multiple recall thresholds. And insteadly, they may over-optimize for metrics such as Recall@500 and Recall@1000, which are easier to improve during evolution.

However, when we employ \texttt{gpt-oss-120b} as the agent’s core LLM, the self-evolution framework consistently outperforms all baseline methods across all recall cutoffs.
It surpasses the greedy strategy by +0.39\%\textasciitilde +1.04\% on the recall scores, as well as an average recall improvement of +0.61\%.

Comparing the performance of \textsc{Evolve} and \textsc{Human-3}, we further find that when using \texttt{gpt-oss-20b} and \texttt{gpt-oss-120b} as the core LLMs, the rule combinations explored by the agent outperform the manually designed rules on all recall metrics, with average improvements of +0.82\% and +1.03\%, respectively. 
These two large LLMs may have strong capabilities, enabling them to follow the procedures defined in the self-evolution framework. They may generate better rules and explore more effective rule combinations.

\section{Discussions}

In this section, we dive into the underlying mechanisms that drive self-evolution
We investigate three research questions: (1) how the quality of individual rules evolves during exploration; (2) how the capacity of the core LLM influences experimental planning when exploiting previously validated rule combinations; and (3) whether the agent can reliably perform conservative, evidence-driven rule elimination. 
{Through these analyses, we aim to shed light on the behavioral characteristics and limitations of self-evolving LLM agents beyond aggregate retrieval metrics.}

\subsection{RQ1: Evolution of Rule Quality?}

The success of \textsc{Greedy} motivates us to examine whether the agent can propose better rules as the self-evolution process conducts.
We analyze the performance of each individual rule and record the time of its creation. 
For each run, the agent evolves for 500 steps, and we partition the process into five phases of 100 steps each.

\begin{figure}[t] 
\centering 
\begin{subfigure}{1\columnwidth}
\resizebox{1\columnwidth}{!}{  
    \includegraphics[width=1\columnwidth]{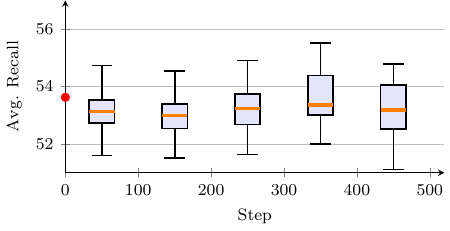}
}
\subcaption{Rules created by \texttt{Qwen3-4B-Thinking}.}
\label{fig:individual_vs_pos_qwen3} 
\end{subfigure}
\vskip 0.25in

\begin{subfigure}{1\columnwidth}
\resizebox{1\columnwidth}{!}{  
    \includegraphics[width=1\columnwidth]{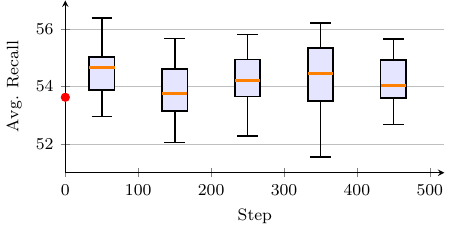}
}
\subcaption{Rules created by \texttt{gpt-oss-120b}.}
\end{subfigure}

\caption{Performance on dev set of individual rules created by self-evolving agents at distinct phases. The red dot at $t=0$ shows the performance of a human-written rule, which is provided as the seed rule for agents.} 
\label{fig:individual_vs_pos}  
\end{figure}

We study the highest- and lowest-performing self-evolving agents for analysis, which employ \texttt{gpt-oss-120b} and \texttt{Qwen3-4B-Thinking} as the core LLM, respectively.
Figure~\ref{fig:individual_vs_pos} illustrates the distribution of rule performance within each phase by boxplot.
Across the two agents, we find that they fail to consistently generate increasingly effective rules over time.
Instead, at each phase of evolution, the agent produces a mixture of both effective and ineffective rules.
Both the median and the maximum recall of newly created rules fluctuate across the five phases, without monotonicity.

The results suggest that the self-evolution framework does not primarily rely on progressive improvement of new rules. 
The process of rule generation may introduce both promising and unproductive rules.
We thus highlight the importance to study downstream mechanisms, such as how the agents with distinct core LLMs plan experiments and eliminate ineffective rules.

\subsection{RQ2: Effect of Core LLM Capacity on Experimental Planning}

We investigate how the agents plan experiments by exploiting previously validated rule combinations.
Different from rule generation, experiment planning requires the agent to reason over the results in $S^{(t)}$, identify promising combinations, and design new experiments that extend previous successes rather than exploring unproductive regions of the search space.

\begin{figure}[t] 
\centering 
\begin{subfigure}{1\columnwidth}
\resizebox{1\columnwidth}{!}{  
    \includegraphics[width=1\columnwidth]{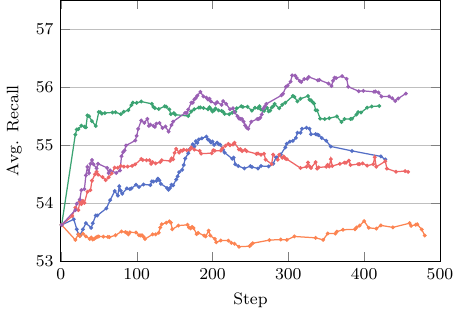}
}
\subcaption{Using \texttt{Qwen3-4B-Thinking} as the core LLM.}
\label{fig:prior_comb_vs_pos_qwen3}
\end{subfigure}
\vskip 0.4in

\begin{subfigure}{1\columnwidth}
\resizebox{1\columnwidth}{!}{  
    \includegraphics[width=1\columnwidth]{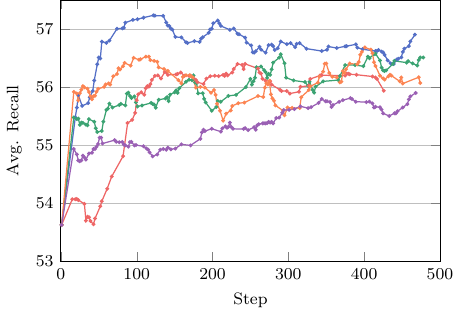}
}
\subcaption{Using \texttt{gpt-oss-120b} as the core LLM.}
\end{subfigure}

\caption{Performance of the rule combinations mentioned by the agent when planning new experiments. We illustrate the curves for five runs respectively.} 
\label{fig:prior_comb_vs_pos}  
\end{figure}

To analyze this process, we examine the chain-of-thought texts~\cite{wei2022chain} generated by the core LLM when the agent plans each new experiment.
We find that the agent frequently justifies its choice of a new rule combination by explicitly mentioning one or more previously validated combinations.
The new experiment is typically framed as incremental modifications of these prior combinations, such as adding or removing one rule. See more case studies in Appendix~\ref{app:exp_plan_case_study}.

We extract the validated combinations referenced by the agents when planning experiments. 
We then compare these \textit{anchor} combinations and record the one with the best performance at each step. 
We track the recall scores of referenced combinations over time in the evolution process of \texttt{Qwen3-4B-Thinking} and \texttt{gpt-oss-120b}.

Figure~\ref{fig:prior_comb_vs_pos} shows the performances of  anchor combinations in five runs. 
We find in several runs of \texttt{Qwen3-4B-Thinking}, the agent fails to select progressively better combinations of rules as references. 
Specifically, in the fifth run (orange curve in Figure~\ref{fig:prior_comb_vs_pos_qwen3}), the agent continuously plans new experiments based on those previous combinations worse than the seed rule. 
However, the results in Figure~\ref{fig:individual_vs_pos_qwen3} shows that \texttt{Qwen3-4B-Thinking} is capable to generate superior individual rules than the seed. 
It indicates that this model's reasoning capabilities may be insufficient to leverage the successful results in the history $S^{(t)}$ for the design of new experiments.
As a result, the agent struggles to explore combinations with better performance and yields suboptimal recalls to the greedy strategy.

In contrast, across all five runs, \texttt{gpt-oss-120b} can select stronger validated combinations as the basis for new experiments, as it self-evolves on more steps. 
We also find that, across multiple runs, \texttt{gpt-oss-120b} can achieve more convergent recall scores at the final stage compared to other LLMs.
We believe that the capabilities to maintain and exploit high-quality planning anchors is a key factor underlying why the agent using \texttt{gpt-oss-120b} is the only one that consistently outperforms the greedy strategy across all recall cutoffs.

\subsection{RQ3: Faithfulness in Rule Elimination}

Rule elimination is a critical component of the self-evolution framework, since premature removal of potentially effective rules may constrain the search space, while permissive retention may lead to ineffective exploration.

\begin{figure}[t] 
\centering 
\begin{subfigure}{1\columnwidth}
\resizebox{1\columnwidth}{!}{  
    \includegraphics[width=1\columnwidth]{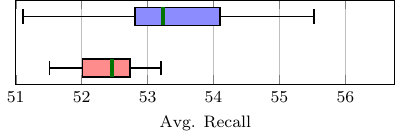}
}
\subcaption{Rules created by \texttt{Qwen3-4B-Thinking}.}
\end{subfigure}
\vskip 0.4in

\begin{subfigure}{1\columnwidth}
\resizebox{1\columnwidth}{!}{  
    \includegraphics[width=1\columnwidth]{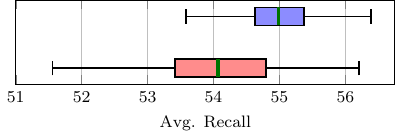}
}
\subcaption{Rules created by \texttt{gpt-oss-120b}.}
\end{subfigure}

\caption{Distributions of the retained rules and the eliminated rules at the end of self-evolution. We employ blue boxes to present the \blue{retained rules}, while red boxes for the \red{eliminated rules}.} 
\label{fig:stat_elim_and_retain}  
\end{figure}

We examine which rules are retained or discarded in the evolution process. Figure~\ref{fig:stat_elim_and_retain} shows the performance distributions of the retained rules and the eliminated rules after self-evolution. 
Compared with the discarded rules, the retained rules have higher median and higher maximum recalls.
The self-evolving agents are capable of discerning the quality of rules and attempt to eliminate ineffective ones.
However, we find that \texttt{gpt-oss-120b} may eliminate several effective rules, which almost achieve the best performance of the retained rules.

We further examine the number of rules retained at the end of evolution. 
When using \texttt{Qwen3-4B-Thinking} as the core LLM, the agent retains an average of 37.8$\pm$3.9 rules, although the prompts in Appendix~\ref{app:prompt_rule_elimination} instruct it to keep the scale of active rules not significantly larger than 6 rules.
Consequently, the search space for rule combinations grows exponentially, preventing \texttt{Qwen3-4B-Thinking} from identifying effective combinations.
\texttt{gpt-oss-120b} can consistently control the number of rules in memory $R^{(t)}$ at 5.8$\pm$1.0 rules, but it sometimes discards a potentially  effective rule without sufficient experiments.

When exposed to lengthy evolving principles, current LLMs  usually overemphasize a limited subset of these principles. For instance, \texttt{gpt-oss-120b} tends to focus on control the number of active rules, which \texttt{Qwen3-4B-Thinking} concentrates more on not eliminating the potentially promising rules. As a result, during the self-evolution process, even under explicit and detailed decision-making constraints, the agents still fail to align with human expert decisions.

\section{Conclusion}

We presented a self-evolution framework that enables an agent to refine query-rewriting rules for legal case retrieval in a training-free setting. 
By interacting with an automated evaluation environment, the agent iteratively generates new rules, plans experiments to assess rule combinations, and eliminates ineffective strategies. 

Experimental results on the LeCaRD-v2 benchmark demonstrate that the proposed framework, when employing a capable core LLM such as \texttt{gpt-oss-120b}, can consistently outperform non-evolutionary baselines. 
Our work highlights the potential of self-evolving agents for interpretable and adaptive enhancement for legal case retrieval.

\section{Limitations}

Legal case retrieval is a critical task that supports real-world practices such as judicial decision-making and legal consultation.
It has been widely adopted across jurisdictions that differ in language and legal tradition. 
Since the structure of case documents, the terminology, and other characteristics may be different across judicial systems, we note that when applying the proposed method to other legal case retrieval benchmarks, it would be better to provide the agent with appropriate seed rules at the initialization stage of self-evolution.

We also note that the core LLM's capabilities to follow instructions play a vital role in self-evolution. We have tried to employ non-reasoning models to serve as the core LLM, for instance, \texttt{Qwen3-30B-A3B-Instruct}~\cite{yang2025qwen3technicalreport}. However, this LLM cannot understand the format constraints in the prompts in Appendix~\ref{app:prompts}, merely generating meaningless rules. For instance, in the description of the rewriting process, the model fails to provide any descriptive texts but merely enumerates several semantically redundant keywords, as shown in Figure~\ref{fig:example_of_instruct_failure}. Thus, we employ four reasoning LLMs in this work, rather than further exploring whether there are non-reasoning LLMs which can serve as the core model in self-evolution.

\begin{figure}[h]
    \centering
    \begin{tcolorbox}[arc=1mm,boxrule=1pt,left=4pt,right=4pt,top=4pt,bottom=4pt]
    \small
    \textit{The original rule in Chinese:}
    
    \zhsmall{[诉讼时效届满]、[时效经过]、[超过诉讼时效]、[时效已过]、[时效失效]、[诉讼时效已过]}

    \zhsmall{格式：[术语1]、[术语2]、[术语3]、[术语4]、[术语5]、[术语6]}
    \end{tcolorbox}
    
    \begin{tcolorbox}[arc=1mm,boxrule=1pt,left=4pt,right=4pt,top=4pt,bottom=4pt]
    \small
    \textit{The translated rule in English:}
    
    [expiration of the statute of limitations], [lapse of the limitation period], [exceeding the statute of limitations], [limitation period has passed], [limitation invalid], [statute of limitations has passed]

    \textbf{Format:} [Term A], [Term B], [Term C], [Term D], [Term E], [Term F]
    \end{tcolorbox}
    \caption{An example of the meaningless rule generated by \texttt{Qwen3-30B-A3B-Instruct}.}
    \label{fig:example_of_instruct_failure}
\end{figure}


We also note that the core LLM's capabilities to understand and to generate Chinese texts are important. We have tried to employ \texttt{Magistral-Small-2509}~\cite{mistralai2025magistral}, a multi-lingual reasoning LLM with 24 billion parameters, as the core model. However, the model usually provides rewriting rules with language-switching and fragmented phrases, as shown in Figure~\ref{fig:example_of_magistral}.
We do not further investigate how LLM's language-specific capabilities affect its performance in self-evolving. Instead, we empirically employ the four LLMs with promising capabilities to understand Chinese instructions and generate Chinese texts.

\begin{figure}[h]
    \centering
    \begin{tcolorbox}[arc=1mm,boxrule=1pt,left=4pt,right=4pt,top=4pt,bottom=4pt]
    \small
    \textit{The original rule in Chinese:}
    
    \zhsmall{扩展性语义域规则：将原始理解后的转换等效性基准，以同义复述成定义标准化。\red{若需区分初中类目式，可覆盖近Uri和底层}}

    \zhsmall{格式：\red{[拓扞符ca \hindiworda][共现][规范整合][metaJesus shakes \koreanwordb 的合规变体]}}
    \end{tcolorbox}
    
    \begin{tcolorbox}[arc=1mm,boxrule=1pt,left=4pt,right=4pt,top=4pt,bottom=4pt]
    \small
    \textit{The translated rule in English:}
    
    \textbf{Extensibility semantic-domain rule:} Transform the equivalence baseline obtained from the original interpretation into a standardized definition through synonymous paraphrasing. \red{If need to distinguish Middle School Kind Formula, it can cover Near Uri and Bottom}

    \textbf{Format:} \red{[Printmaking Symbol ca \hindiworda], [Co-occurrence], [Standard Combination], [metaJesus shakes \koreanwordb's Legal Varients]}
    \end{tcolorbox}
    \caption{An example of the fragmented rule generated by \texttt{Magistral-Small-2509}.}
    \label{fig:example_of_magistral}
\end{figure}

\section*{Acknowledgments}
This work is supported in part by Self-Determined Project of the Discipline and Technology Research Center for Large Model Intelligence Applications. We thank the anonymous reviewers for their helpful discussions and suggestions.

\bibliography{latex/custom}

\newpage

\appendix

\section{Prompts Employed in the Self-Evolution Framework}
\label{app:prompts}

This section details the prompts employed in our self-evolution framework, including those for action decision, rule generation, experiment planning, and the elimination of ineffective rules.

At each step $t$ of the self-evolution process, the agent need to make a decision based on its memory, i.e., the active rule set $R^{(t)}$, the eliminated rule set $\bar{R}^{(t)}$ and the experimental history $S^{(t)}$. 
We use the mark \texttt{\$\{rules\}} to denote the context representing $R^{(t)}$, while the mark \texttt{\$\{elim\_rules\}} for $\bar{R}^{(t)}$ and the mark \texttt{\$\{scores\}} for $S^{(t)}$.

We also provide the performance of vanilla BM25 to the agent, whose context is inserted into the slot \texttt{\$\{baseline\_scores\}}.

\subsection{Prompts of Action Decision}
\label{app:prompt_action_decision}

We present the Chinese prompt in Appendix~\ref{app:prompt_action_chinese} and the corresponding English translated prompt in Appendix~\ref{app:prompt_action_english}.

\subsubsection{The Original Chinese Prompt of Action Decision}
\label{app:prompt_action_chinese}
\begin{promptbox}{The original prompt of action decision}

\zhsmall{\ \ \ \ \ \ 你是一名资深的法律专家。你的任务是构造高效的查询改写规则集合，提升法律文档检索的效果。你需要不断调整现有的检索查询改写规则，在此过程中，你可以采取**引入新的规则**、**开展实验验证规则的有效性**、**剔除低效规则**等方式。现在，你需要对接下来的工作进行规划，决定下一步应该优先采取哪一种方式。}

\medskip

\zhsmall{\# 现有的查询改写规则：}

\texttt{\$\{rules\}}

\medskip

\zhsmall{\# 未进行改写时的结果：}

\texttt{\$\{baseline\_score\}}

\medskip

\zhsmall{\# 利用已有改写规则的排列组合开展实验的结果：}

\texttt{\$\{scores\}}

\medskip

\zhsmall{\# 决策历史：}

\texttt{\$\{action\_history\}}

\medskip

\zhsmall{\# 要求：}

\zhsmall{\ \ \ \ \ \ \ \ 在决策前，你可以参考至多向前10步（即T-10\\ 时刻，当前为T-0时刻）的决策历史，并从中吸取教训，避免频繁失败。请你结合现有的查询改写规则及上述实验结果，选择下一步的动作。你可以选择**开展实验验证**、**剔除低效规则**、**创造新的规则**三种方式中的一种。}

\medskip

\zhsmall{\ \ \ \ \ \ \ \ 低效规则是指：根据该规则进行查询改写之\\ 后，检索的结果不仅没有得到改进，其召回率反而低于不使用该规则的实验。请注意，在判断某条规则是否为低效规则时，你应当非常小心，尽管引入某条规则在部分实验中可能导致检索性能下降，但在另一些实验中，引入这些规则可能有助于提高检索的召回率，这些规则不应被判定为低效规则。同时，请你尽量将现有的查询改写规则控制在6条之内，如果现有规则超过上限，此时某条改写规则A在单独使用或与任意其他规则组合时带来的提升均小于另一条改写规则B，则也可以将改写规则A视作低效规则。}

\medskip

\zhsmall{你在选择下一步动作时，应遵循如下原则：}

\zhsmall{\ \ \ - 如果现有规则的排列组合还未得到充分验证，此时应当选取某一条规则或者某几条规则的集合，测试按照这些规则改写查询语的效果，你应优先选择**开展实验验证**；}

\zhsmall{\ \ \ - 如果你认为存在部分规则明显是低效的、会导致改写后的检索效果变差，你可以选择**剔除低效规则**；}

\zhsmall{\ \ \ - 如果基于现有规则的排列组合基本都已得到验证，且没有明显的低效规则需要剔除，你可以选择**创造新的规则**。}

\medskip

\zhsmall{\ \ \ \ \ \ \ \ 请你按照下列**模板**的形式，先简要说明为什么选择该方式作为下一步的动作，然后明确地输出该动作的名称，你必须从三个备选动作中选取下一步的动作，且只能选取一个。再次强调，请你严格按照**模板**中的形式分段输出，并且严禁提供其他内容或者做任何额外说明。}

\medskip

\zhsmall{\# 模板：}

\zhsmall{**分析**：}

\zhsmall{\{\{一段文本，简要说明为什么选取该动作\}\}}

\zhsmall{**下一步动作**:}

\zhsmall{\{\{开展实验验证/剔除低效规则/创造新的规则\}\}}

\end{promptbox}

\subsubsection{The Translated Prompt of Action Decision}
\label{app:prompt_action_english}
\begin{promptbox}{The translated prompt of action decision}

\zhsmall{\ \ \ \ You are a senior legal expert. Your task is to construct an efficient set of query rewriting rules to improve the effectiveness of legal document retrieval. You need to continuously adjust the existing query rewriting rules. In this process, you may adopt actions such as **introducing a new rule**, **conducting experimental validation of rule effectiveness**, and **removing an ineffective Rule**. Now, you need to decide which action should be prioritized next.}

\medskip

\zhsmall{\# Active Query Rewriting Rules:}

\texttt{\$\{rules\}}

\medskip

\zhsmall{\# Baseline Recall Scores without Rewriting:}

\texttt{\$\{baseline\_score\}}

\medskip

\zhsmall{\# Experimental Results of the Combination of Existing Rules:}

\texttt{\$\{scores\}}

\medskip

\zhsmall{\# Action History:}

\texttt{\$\{action\_history\}}

\medskip

\zhsmall{\# Instructions:}

\zhsmall{\ \ \ \ Before making a decision, you may refer to up to the previous ten steps of decision history (i.e., from time step T-10, with the current step being T-0) and draw lessons from past outcomes to avoid repeated failures. Based on the existing query-rewriting rules and the experimental results described above, please select the next action. You may choose one of the following options: **conduct experimental validation**, **eliminate an ineffective rule**, or **create a new rule**.}

\medskip

\zhsmall{\ \ \ \ Definition of Ineffective Rule: When applying this rule to rewrite a query, the retrieval results are not improved; instead, the recall scores are lower than that of experiments in which the rule is not used. Please note that when determining whether a rule is ineffective, you should be very cautious. Although introducing a rule may result in degraded retrieval performance in some experiments, the same rule may help improve retrieval recall in other experiments; such rules should not be classified as ineffective rules. Meanwhile, please try to keep the total number of existing query rewriting rules within six. If the number of existing rules exceeds this limit, you can regard a rule \textit{A} as an ineffective rule, if the improvement brought by rule \textit{A} is consistently smaller than that brought by another rule \textit{B}, when used alone or in combination with any other rules.}

\medskip

\zhsmall{\ \ \ \ When choosing the next action, you should adhere to the following principles.}

\zhsmall{\ \ - If the combinations of the existing rules have not been sufficiently validated, you should prioritize to **conduct experimental validation**. Specifically, you can select an individual rule or a subset of rules and evaluate the effectiveness of query rewriting under these rules.}

\zhsmall{\ \ - If certain rules are deemed to be evidently ineffective and are likely to degrade retrieval performance after query rewriting, they may be excluded to improve overall effectiveness, and you can choose to **eliminate an ineffective rule**.}

\zhsmall{\ \ - If the combinations of the existing rules have largely been validated and no clearly ineffective rules remain to be eliminated, you can choose to **create a new rule** to further enhance the rewriting strategy.}

\medskip

\zhsmall{\ \ \ \ Please follow the format of the **template** below. First, briefly explain why this action is chosen for the next step, and then explicitly output the name of that action. You must select the next action from three candidate actions, and you can select only one. Emphasize again that you must strictly follow the format in the **template** for segmented output, and you are strictly prohibited from providing any other content or making any additional explanations.}

\medskip

\zhsmall{\# Template:}

\zhsmall{**Analysis**：}

\zhsmall{\{\{A segment of text that briefly explains why this action is selected.\}\}}

\zhsmall{**The Next Action**:}

\zhsmall{\{\{Conduct Experimental Validation / Eliminate an Ineffective Rule / Create a New Rule\}\}}

\end{promptbox}

\subsection{Prompts of Rule Generation}

We present the Chinese prompt in Appendix~\ref{app:prompt_rule_gen_chinese} and the corresponding English translated prompt in Appendix~\ref{app:prompt_rule_gen_english}.

\subsubsection{The Original Chinese Prompt of Rule Generation}
\label{app:prompt_rule_gen_chinese}

\begin{promptbox}{The original prompt of rule generation}

\zhsmall{\ \ \ \ \ \ 你是一名资深的法律专家。你的任务是根据已有的查询改写规则，设计出一个新的查询改写规则，以提升法律文档检索的效果。}

\medskip

\zhsmall{\# 已被证实改写效果较差并被剔除的改写规则：}

\texttt{\$\{elim\_rules\}}

\medskip

\zhsmall{\# 现有的查询改写规则：}

\texttt{\$\{rules\}}

\medskip

\zhsmall{\# 未进行改写时的结果：}

\texttt{\$\{baseline\_score\}}

\medskip

\zhsmall{\# 利用已有改写规则的排列组合开展实验的结果：}

\texttt{\$\{scores\}}

\medskip

\zhsmall{\# 要求：}

\zhsmall{\ \ \ \ \ \ 请你结合现有的查询改写规则及上述实验结果，设计出一个新的查询改写规则。你需要检视那些被剔除的规则，它们很有可能无法带来有效的提升，你在创造新规则时，应避免给出与这些规则较为相似的新规则。你可以对现有规则进行修改，也可以提出全新的思路。请确保新规则能够有效提升法律文档的检索效果。}

\zhsmall{\ \ \ \ \ \ \ 请注意，在执行检索时，会自动将改写后的查询与原始查询进行拼接，因此你设计的新查询规则中**禁止**重复输出原始查询。}

\zhsmall{\ \ \ \ \ \ \ \ 请你只提供一条新的规则，形式仿照**现有的\\ 查询改写规则**，你需要按照下列**模板**的形式，先简要说明该规则的设计思路和预期效果，再明确写出该规则的名称、具体内容、输出格式。再次强调，请你严格按照**模板**中的形式分段输出，并且严禁提供其他内容或者做任何额外说明。}

\medskip

\zhsmall{\# 模板：}

\zhsmall{**设计思路**：}

\zhsmall{\{\{一段文本，简要说明该规则的设计思路\}\}}

\zhsmall{**预期效果**：}

\zhsmall{\{\{一段文本，说明该规则预期提升的检索效果\}\}}

\zhsmall{**新规则内容**:}

\zhsmall{\{\{规则名称\}\}：\{\{一句话或几句话，描述具体的改写方法\}\}}

\zhsmall{格式：\{\{一行文本，顿号间隔的一个或多个使用半角方括号"[]"做标记的格式类型，描述输出文本的格式\}\}}

\end{promptbox}

\subsubsection{The Translated Prompt of Rule Generation}
\label{app:prompt_rule_gen_english}

\begin{promptbox}{The translated prompt of rule generation}

\zhsmall{\ \ \ \ You are a senior legal expert. Your task is to design a new rule for query rewriting based on the existing ones, in order to improve the effectiveness of legal document retrieval.}

\medskip

\zhsmall{\# The eliminated rules that have been proved to be ineffective：}

\texttt{\$\{elim\_rules\}}

\medskip

\zhsmall{\# Active Query Rewriting Rules:}

\texttt{\$\{rules\}}

\medskip

\zhsmall{\# Baseline Recall Scores without Rewriting:}

\texttt{\$\{baseline\_score\}}

\medskip

\zhsmall{\# Experimental Results of the Combination of Existing Rules:}

\texttt{\$\{scores\}}

\medskip

\zhsmall{\# Instructions:}

\zhsmall{\ \ \ \ Please design a new query rewriting rule based on the existing rules and the experimental results provided above. You need to examine those rules that have been discarded, as they are likely unable to bring effective improvements. When creating new rules, you should avoid proposing rules that are highly similar to these discarded ones. You may modify existing rules or propose entirely new ideas. Please ensure that the new rule can effectively improve the retrieval performance of legal documents.}

\zhsmall{\ \ \ \ Please note that during retrieval, the rewritten query will be automatically concatenated with the original query. Therefore, your are **prohibited** to provide a new rule using the repeated inclusion of the original query.}

\zhsmall{\ \ \ \ Please provide only one new rule, following the format of the **active query rewriting rules**. You need to first briefly explain the design rationale and expected effect of the rule according to the **template** below. Then clearly specify the rule’s name, detailed content, and output formatting.
Emphasize again that you must strictly follow the format in the **template** for segmented output, and you are strictly prohibited from providing any other content or making any additional explanations.}

\medskip

\zhsmall{\# Template:}

\zhsmall{**Design Concept**:}

\zhsmall{\{\{A brief description outlining the design rationale behind this rule.\}\}}

\zhsmall{**Expected Outcome**:}

\zhsmall{\{\{A brief description explaining the anticipated improvement in retrieval performance resulting from this rule.\}\}}

\zhsmall{**New Rule**:}

\zhsmall{\{\{Rule Name\}\}: \{\{one or more sentences describing the process of query rewriting.\}\}}

\zhsmall{Formatting: \{\{A single line of text
A single line of text containing one or more format types enclosed in half-width square brackets "[]" and separated by commas, describing the format of the rewritten query.\}\}}

\end{promptbox}

\subsection{Prompts of Experiment Planning}

We present the Chinese prompt in Appendix~\ref{app:prompt_exp_plan_chinese} and the corresponding English translated prompt in Appendix~\ref{app:prompt_exp_plan_english}.

\subsubsection{The Original Chinese Prompt of Experiment Planning}
\label{app:prompt_exp_plan_chinese}

\begin{promptbox}{The original prompt of experiment planning}

\zhsmall{\ \ \ \ \ \ 你是一名资深的法律专家。你的任务是构造高效的查询改写规则集合，提升法律文档检索的效果。你需要不断调整现有的检索查询改写规则，在此过程中，你可能需要开展实验，验证规则组合的有效性。}

\medskip

\zhsmall{\# 现有的查询改写规则：}

\texttt{\$\{rules\}}

\medskip

\zhsmall{\# 未进行改写时的结果：}

\texttt{\$\{baseline\_score\}}

\medskip

\zhsmall{\# 利用已有改写规则的排列组合开展实验的结果：}

\texttt{\$\{scores\}}

\medskip

\zhsmall{\# 要求：}

\zhsmall{\ \ \ \ \ \ \ \ 请你结合现有的查询改写规则及上述实验结\\ 果，决定下一步应利用哪些现有规则开展实验，以验证其有效性。你需要按照下列**模板**的形式，先简要说明为什么选择这些规则进行实验验证，然后明确地输出这些规则的编号列表，编号之间用逗号分隔。注意，你可以选择一条规则，也可以选择多条规则的组合进行实验验证，但你目前仅能开展一次实验，你只可以输出一组规则编号列表。再次强调，请你严格按照**模板**中的形式分段输出，并且严禁提供其他内容或者做任何额外说明。}

\medskip

\zhsmall{\# 模板：}

\zhsmall{**分析**：}

\zhsmall{\{\{一段文本，简要说明为什么选取这些规则进行实验\}\}}

\zhsmall{**下一步实验需使用的规则编号列表**:}

\zhsmall{\{\{1/2/3/1,2/1,3/2,3/1,2,3/……\}\}}

\end{promptbox}

\subsubsection{The Translated Prompt of Experiment Planning}
\label{app:prompt_exp_plan_english}

\begin{promptbox}{The translated prompt of experiment planning}

\zhsmall{\ \ \ \ You are a senior legal expert. Your task is to construct an efficient set of query rewriting rules to improve the effectiveness of legal document retrieval. You need to continuously adjust the existing query rewriting rules. In this process, you may need to conduct experiments to verify the effectiveness of the rule combinations.}

\medskip

\zhsmall{\# Active Query Rewriting Rules:}

\texttt{\$\{rules\}}

\medskip

\zhsmall{\# Baseline Recall Scores without Rewriting:}

\texttt{\$\{baseline\_score\}}

\medskip

\zhsmall{\# Experimental Results of the Combination of Existing Rules:}

\texttt{\$\{scores\}}

\medskip

\zhsmall{\# Instructions:}

\zhsmall{\ \ \ \ Please select a combination of rules to validate its effectiveness based on the existing rules and the experimental results provided above. You need to follow the template below to briefly explain why these rules are chosen for experimental validation, then clearly output the list of rule numbers, separated by commas. Note that you may select one single rule or one combination of multiple rules for experimental validation. Since you can only conduct one experiment at present, so you may only output one set of rule numbers. Emphasize again that you must strictly follow the format in the **template** for segmented output, and you are strictly prohibited from providing any other content or making any additional explanations.}

\medskip

\zhsmall{\# Template:}

\zhsmall{**Analysis**:}

\zhsmall{\{\{A brief explanation of the rationale behind selecting these rules for the experiment.\}\}}

\zhsmall{**List of rule numbers needed for the next step experiment**:}

\zhsmall{\{\{1/2/3/1,2/1,3/2,3/1,2,3/……\}\}}

\end{promptbox}

\subsection{Prompts of Rule Elimination}
\label{app:prompt_rule_elimination}

In the step of rule elimination, the agent should first determine whether there is an ineffective rule in $R$. We present the original prompt in Appendix~\ref{app:rule_elim_judge_chinese} and the translated prompt in Appendix~\ref{app:rule_elim_judge_ebglish}.

If the agent confirm the necessity to eliminate a rule, it should then determine which rule should be removed. We present the original prompt in Appendix~\ref{app:select_elim_rule_chinese} and the translated prompt in Appendix~\ref{app:select_elim_rule_english}.

\subsubsection{The Original Prompts of Determining Whether to Proceed Elimination}
\label{app:rule_elim_judge_chinese}
\begin{promptbox}{The original prompt of determining the necessity of elimination}

\zhsmall{\ \ \ \ \ \ 你是一名资深的法律专家。你的任务是分析已有的查询改写规则中是否存在可能导致检索效果变差的规则。}

\medskip

\zhsmall{\# 现有的查询改写规则：}

\texttt{\$\{rules\}}

\medskip

\zhsmall{\# 未进行改写时的结果：}

\texttt{\$\{baseline\_score\}}

\medskip

\zhsmall{\# 利用已有改写规则的排列组合开展实验的结果：}

\texttt{\$\{scores\}}

\medskip

\zhsmall{\# 要求：}

\zhsmall{\ \ \ \ \ \ 请你结合现有的查询改写规则及上述实验结果，分析是否存在某条低效规则。}

\medskip

\zhsmall{\ \ \ \ \ \ \ \ 低效规则是指：根据该规则进行查询改写之\\ 后，检索的结果不仅没有得到改进，其召回率反而低于不使用该规则的实验。请注意，在判断某条规则是否为低效规则时，你应当非常小心，尽管引入某条规则在部分实验中可能导致检索性能下降，但在另一些实验中，引入这些规则可能有助于提高检索的召回率，这些规则不应被判定为低效规则。同时，请你尽量将现有的查询改写规则控制在6条之内，如果现有规则超过上限，此时某条改写规则A在单独使用或与任意其他规则组合时带来的提升均小于另一条改写规则B，则也可以将改写规则A视作低效规则。}

\medskip

\zhsmall{\ \ \ \ \ \ 请你谨慎的判断是否存在需要被剔除的改写规则，你需要按照下列**模板**的形式，给出明确的理由说明为什么存在被剔除的规则，再下最终的结论，你的结论只能是“存在需要被剔除的规则”或“不存在需要被剔除的规则”二者之一。再次强调，请你严格按照**模板**中的形式分段输出，并且严禁提供其他内容或者做任何额外说明。}

\medskip

\zhsmall{\# 模板：}

\zhsmall{**分析**：}

\zhsmall{\{\{一段文本，简要说明为什么存在需要被剔除的改写规则\}\}}

\zhsmall{**结论**:}

\zhsmall{\{\{存在需要被剔除的规则/不存在需要被剔除的规则\}\}}

\end{promptbox}

\subsubsection{The Original Prompt of Selecting a Rule to Be Eliminated}
\label{app:select_elim_rule_chinese}
\begin{promptbox}{The original prompt of determining which rule to be eliminated}

\zhsmall{\ \ \ \ \ \ 你是一名资深的法律专家。你的任务是分析已有的查询改写规则中，哪条规则的存在最有可能导致检索效果变差，需要优先被剔除。}

\medskip

\zhsmall{\# 现有的查询改写规则：}

\texttt{\$\{rules\}}

\medskip

\zhsmall{\# 未进行改写时的结果：}

\texttt{\$\{baseline\_score\}}

\medskip

\zhsmall{\# 利用已有改写规则的排列组合开展实验的结果：}

\texttt{\$\{scores\}}

\medskip

\zhsmall{\# 要求：}

\zhsmall{\ \ \ \ \ \ 请你结合现有的查询改写规则及上述实验结果，分析是否存在某条低效规则。}

\medskip

\zhsmall{\ \ \ \ \ \ \ \ 低效规则是指：根据该规则进行查询改写之\\ 后，检索的结果不仅没有得到改进，其召回率反而低于不使用该规则的实验。请注意，在判断某条规则是否为低效规则时，你应当非常小心，尽管引入某条规则在部分实验中可能导致检索性能下降，但在另一些实验中，引入这些规则可能有助于提高检索的召回率，这些规则不应被判定为低效规则。同时，请你尽量将现有的查询改写规则控制在6条之内，如果现有规则超过上限，此时某条改写规则A在单独使用或与任意其他规则组合时带来的提升均小于另一条改写规则B，则也可以将改写规则A视作低效规则。}

\medskip

\zhsmall{\ \ \ \ \ \ 请你选择一条最应该被优先剔除的规则，你需要按照下列**模板**的形式，给出明确的理由说明为什么根据该条规则改写之后检索的效果会变差，以及为什么这条规则需要先于其他规则被剔除，再明确地输出该规则的编号。注意，你必需输出且仅能输出**1个**待剔除规则的编号，你输出的编号应为纯数字形式，不要输出规则的名称、具体内容、改写格式等信息。再次强调，请你严格按照**模板**中的形式分段输出，并且严禁提供其他内容或者做任何额外说明。}

\medskip

\zhsmall{\# 模板：}

\zhsmall{**分析**：}

\zhsmall{\{\{一段文本，简要说明为什么存在需要被剔除的改写规则\}\}}

\zhsmall{**应优先被剔除的规则编号**:}

\zhsmall{\{\{1/2/3/……\}\}}

\end{promptbox}

\subsubsection{The Translated Prompt of Determining Whether to Proceed Elimination}
\label{app:rule_elim_judge_ebglish}
\begin{promptbox}{The translated prompt of determining the necessity of elimination}

\zhsmall{\ \ \ \ You are a senior legal expert. Your task is to analyse whether there exists one or more rules which may result in a deteriorated performance after query rewriting.}

\medskip

\zhsmall{\# Active Query Rewriting Rules:}

\texttt{\$\{rules\}}

\medskip

\zhsmall{\# Baseline Recall Scores without Rewriting:}

\texttt{\$\{baseline\_score\}}

\medskip

\zhsmall{\# Experimental Results of the Combination of Existing Rules:}

\texttt{\$\{scores\}}

\medskip

\zhsmall{\# Instructions:}

\zhsmall{\ \ \ \ Based on the existing query rewriting rules and the experimental results above, please analyse whether any ineffective rule is present.}

\medskip

\zhsmall{\ \ \ \ Definition of Ineffective Rule: When applying this rule to rewrite a query, the retrieval results are not improved; instead, the recall scores are lower than that of experiments in which the rule is not used. Please note that when determining whether a rule is ineffective, you should be very cautious. Although introducing a rule may result in degraded retrieval performance in some experiments, the same rule may help improve retrieval recall in other experiments; such rules should not be classified as ineffective rules. Meanwhile, please try to keep the total number of existing query rewriting rules within six. If the number of existing rules exceeds this limit, you can regard a rule A as an ineffective rule, if the improvement brought by rule A is consistently smaller than that brought by another rule B, when used alone or in combination with any other rules.}

\medskip

\zhsmall{\ \ \ \ Please carefully judge whether there exist rewriting rules that need to be removed. You need to provide a clear explanation according to the following **template** on why such rules should be removed, and then give the final conclusion. Your conclusion can only be one of the following two: "There exist rewriting rules that need to be removed" or "There do not exist rewriting rules that need to be removed." Emphasize again that you must strictly follow the format in the **template** for segmented output, and you are strictly prohibited from providing any other content or making any additional explanations.}

\medskip

\zhsmall{\# Template：}

\zhsmall{**Analysis**：}

\zhsmall{\{\{A brief explanation outlining the necessity of eliminating certain rewriting rules.\}\}}

\zhsmall{**Conclusion**:}

\zhsmall{\{\{There exist rewriting rules that need to be removed/There do not exist rewriting rules that need to be removed\}\}}

\end{promptbox}

\subsubsection{The Translated Prompt of Selecting a Rule to Be Eliminated}
\label{app:select_elim_rule_english}
\begin{promptbox}{The translated prompt of determining which rule to be eliminated}

\zhsmall{\ \ \ \ You are a senior legal expert. Your task is to analyse the existing query rewriting rules and identify which rule is most likely to cause a decline in retrieval performance.}

\medskip

\zhsmall{\# Active Query Rewriting Rules:}

\texttt{\$\{rules\}}

\medskip

\zhsmall{\# Baseline Recall Scores without Rewriting:}

\texttt{\$\{baseline\_score\}}

\medskip

\zhsmall{\# Experimental Results of the Combination of Existing Rules:}

\texttt{\$\{scores\}}

\medskip

\zhsmall{\# Instructions:}

\zhsmall{\ \ \ \ Please analyze whether there exists any ineffective rule, based on the existing query rewriting rules and the experimental results.}

\medskip

\zhsmall{\ \ \ \ Definition of Ineffective Rule: When applying this rule to rewrite a query, the retrieval results are not improved; instead, the recall scores are lower than that of experiments in which the rule is not used. Please note that when determining whether a rule is ineffective, you should be very cautious. Although introducing a rule may result in degraded retrieval performance in some experiments, the same rule may help improve retrieval recall in other experiments; such rules should not be classified as ineffective Rules. Meanwhile, please try to keep the total number of existing query rewriting rules within six. If the number of existing rules exceeds this limit, you can regard a rule A as an ineffective rule, if the improvement brought by rule A is consistently smaller than that brought by another rule B, when used alone or in combination with any other rules.}

\medskip

\zhsmall{\ \ \ \ \ \ Please select the single rule that should be prioritized for removal. Using the **template** below, provide a clear explanation of why rewriting according to this rule leads to worse retrieval performance, and why this rule must be removed before any other. Then explicitly output the rule’s number.
Note that you must output **exactly one** rule number for removal. The output should be a pure numeric form only—do not include the rule name, specific content, rewriting format, or any other information. Emphasize again that you must strictly follow the format in the **template** for segmented output, and you are strictly prohibited from providing any other content or making any additional explanations.}

\medskip

\zhsmall{\# Template:}

\zhsmall{**Analysis**：}

\zhsmall{\{\{A brief explanation of why a certain rewriting rule must be removed.\}\}}

\zhsmall{**A rule number that should be removed with priority**:}

\zhsmall{\{\{1/2/3/……\}\}}

\end{promptbox}

\section{Prompts of Query Rewriting}
\label{app:prompt_rewriting}

In this work, we employ \texttt{Qwen3-4B-Thinking} as the rewriting model for all experiments. We use following prompts for query rewriting.

\begin{promptbox}{The original prompt of query rewriting}

\zhsmall{\ \ \ \ \ \ \ 你是一名法律专家，你的最终目标是利用检索器从历史案件数据库中搜集与**待决案件**相似的案件，即待决案件的“类案”。}

\zhsmall{\ \ \ \ \ \ \ 现在，请你结合下列**修改规则**，基于**待决案件基本事实**进行修改。修改后的文本将作为原有检索查询语的增强，以确保待决案件的类案出现在检索结果中。}

\medskip

\zhsmall{\# 修改规则：}

\texttt{\$\{rules\}}

\medskip

\zhsmall{\# 待决案件基本事实：}

\texttt{\$\{query\}}

\medskip

\zhsmall{\# 要求：}

\zhsmall{\ \ \ \ \ \ \ \ 你需要严格遵循**修改规则**，以**待决案件基本事实**为基础，将其修改为更适合作为检索查询语的文本。已知，数据库中存储的是历史案件的全文，包括基本事实、裁判理由、判决结果等内\\ 容。检索器采用的是BM25方法。}

\zhsmall{\ \ \ \ \ \ \ 请注意，你无需重新输出一遍待决案件基本事实的原文。}

\zhsmall{\ \ \ \ \ \ 请你在思考后，针对每一条修改规则，按照该规则的改写策略，并严格遵循规则后附的格式要求，直接输出相应的增强查询语，不要做额外解释。}

\zhsmall{\ \ \ \ \ \ \ 在输出时，请你以“修改后的检索查询语：”作为总体的开头。之后，每条规则对应的增强查询语依次以数字标号1、2、3……作为起始。请不要提供“修改说明”之类的冗余解释。}

\end{promptbox}

\begin{promptbox}{The translated prompt of query rewriting}

\zhsmall{\ \ \ \ You are a legal expert, and your ultimate goal is to use a retriever to collect cases from a historical case database that are similar to the **pending case**.}

\zhsmall{\ \ \ \ Now, please revise the query in accordance with the following rules, based on the basic facts of the pending case.
The rewritten query will be used as an enhancement of the original retrieval query, in order to ensure that cases similar to the pending case appear in the retrieval results.}

\medskip

\zhsmall{\# Rewriting Rules:}

\texttt{\$\{rules\}}

\medskip

\zhsmall{\# Facts of the Pending Case:}

\texttt{\$\{query\}}

\medskip

\zhsmall{\# Instructions：}

\zhsmall{\ \ \ \ You need to strictly follow the modification rules and, based on the basic facts of the pending case, modify it into text that is more suitable for use as a retrieval query. It is known that the database stores the full text of historical cases, including basic facts, reasoning of the judgment, judgment results, and other content. The retriever adopts the BM25 method.}

\zhsmall{\ \ \ \ Please note that you do not need to reproduce the original text of the basic facts of the pending case.}

\zhsmall{\ \ \ \ After thinking carefully, for each rule, rewrite according to the rewriting strategy of that rule, and strictly follow the format requirements attached to the rule, directly output the corresponding enhanced query, without providing any additional explanation.}

\zhsmall{\ \ \ \ In the output, begin with the heading “Revised Retrieval Queries:”. Then, present the enhanced query corresponding to each rule sequentially, starting with numerical labels 1, 2, 3, and so on. Do not include any redundant explanations such as “modification notes.”}

\end{promptbox}

\section{More Implementation Details}
\label{app:imple_details}

Following previous work~\cite{levardv2-li-et-al-2024}, we employ Pyserini Lucene~\cite{Lin_etal_SIGIR2021_Pyserini} to implement BM25 retrieval. The judicial document of a case typically comprises three core components: the facts, the legal bases (i.e., statutes and legal reasons), and the judicial decision results.

To align real-world practice of judicial decision, we use the full texts of documents to construct the index, while using only the section of facts as the query during retrieval.

During self-evolution and for all query writing experiments, we do not adjust the hyper-parameters of LLMs for generation. We adopt the default temperatures, top-p, top-k, and other hyper-parameters.

\section{Case Study for Experiment Planning}
\label{app:exp_plan_case_study}

We study how the agent plan the next experiment based on historical results via reasoning. Figure~\ref{fig:case_plan} shows a case generated by \texttt{gpt-oss-120b} when selecting a rule combination to validate.

We find the agent can explicitly mention the rule combinations which have been validated at previous steps. From the analysis provided by LLM, we can identify two reference combinations, the subset $\{2, 3\}$ and the subset $\{1, 2, 3\}$.
The agent finally chooses to exmaine the subset $\{1, 2, 3, 6\}$, we thus denote the mentioned combination with minimum edit distance~(i.e., the subset $\{1,2,3\}$) as the \textit{anchor} experiment for planning.

\begin{figure}[h]
    \centering
    \begin{tcolorbox}[arc=1mm,boxrule=1pt,left=4pt,right=4pt,top=4pt,bottom=4pt,colback=yellow!20!white]
    \textit{The original outputs:}
    
    \zhsmall{\ \ \ \ **分析**：}
    
    \zhsmall{\ \ \ \ 当前实验中，规则组合[2, 3] 已取得最高的 Recall@1K（79.4509），且加入规则 1 后（组合 [1, 2, 3]）进一步提升至 79.6803，显示规则 1、2、3 的协同增益明显。规则 6 单独效果一般，但有潜力在已有高效组合上提供额外的检索线索，尚未与 [1, 2, 3] 组合实验。为验证是否能够进一步突破当前最高召回，计划在已证明有效的 [1, 2, 3] 基础上加入规则 6，进行一次实验。}
    
    \zhsmall{\ \ \ \ **下一步实验需使用的规则编号列表**:}
    
    \zhsmall{\ \ \ \ 1,2,3,6}
    \end{tcolorbox}
    
    \begin{tcolorbox}[arc=1mm,boxrule=1pt,left=4pt,right=4pt,top=4pt,bottom=4pt,colback=yellow!20!white]
    \small
    \textit{The translated outputs:}
    
    \ \ \ \ **Analysis**:
    
    \ \ \ \ In the existing experiments, the rule combination [2, 3] achieves one of the highest Recall@1K (79.4509). After incorporating Rule 1 (i.e., combination [1, 2, 3]), the performance is further improved to 79.6803, indicating a clear synergistic gain among Rules 1, 2, and 3. Rule 6 alone shows moderate effectiveness, but it has the potential to provide additional retrieval cues on top of existing high-performing combinations and has not yet been evaluated together with [1, 2, 3]. To verify whether the current best recall can be further improved, we plan to conduct an experiment by adding Rule 6 to the validated combination [1, 2, 3].

    \ \ \ \ **List of rule numbers needed for the next step experiment**:
    
    \ \ \ \ 1,2,3,6
    \end{tcolorbox}
    
    \caption{An case of the reasoning process for experiment planning, with \texttt{gpt-oss-120b} as the core LLM.}
    \label{fig:case_plan}
\end{figure}

\end{document}